\pdfoutput=1

\documentclass[11pt]{article}

\usepackage{acl}

\usepackage{times}
\usepackage{latexsym}
\usepackage[T1]{fontenc} 
\usepackage[utf8]{inputenc}
\usepackage{microtype} 
\usepackage{inconsolata}
\usepackage{graphicx} 

\usepackage{amsmath} 
\usepackage{booktabs} 
\usepackage{multirow}
\usepackage{colortbl}
\usepackage{xcolor}
\usepackage{graphicx}
\usepackage{natbib} 
\usepackage{subcaption}
\usepackage{enumitem}

\title{DOSE: Data Selection for Multi-Modal LLMs via Off-the-Shelf Models}

\author{
  \textbf{  Biao Wu$^{1}$, Yiwu Zhong$^{2}$, Meng Fang$^{3}$, Ling Chen$^{1}$} \\
  $^1$Australian Artificial Intelligence Institute\\ 
  $^2$Peking University, $^3$University of Liverpool \\ 
  \texttt{biao.wu-2@student.uts.edu.au,zyw@pku.edu.cn}  \\ 
  \texttt{Meng.Fang@liverpool.ac.uk,Ling.Chen@uts.edu.au} 
}

\begin{document}
\maketitle
\begin{abstract}

 High-quality and diverse multimodal data are essential for improving vision–language models (VLMs), yet existing datasets often contain noisy, redundant, and poorly aligned samples. To address these problems, data filtering is commonly used to enhance the efficiency and performance of multimodal learning, but it introduces extra computational cost because filtering models are usually trained on the same data they are meant to screen. To reduce this cost, we study DOSE, which explores whether off-the-shelf pretrained models that have never seen the target data can be used to select training samples for larger and stronger multimodal models without any task-specific training. Even without fine-tuning, these models can effectively assess text quality and image–text alignment to guide data selection. Based on this, we build a joint quality–alignment distribution and apply adaptive weighted sampling to select informative samples while maintaining long-tail diversity. This approach enhances data diversity, enabling models trained on DOSE-filtered data to match or surpass those trained on the full dataset on standard VQA and math benchmarks. Extensive experiments demonstrate its effectiveness, efficiency, and scalability.

\end{abstract}

\vspace{2mm}
\section{Introduction}

\begin{figure}[tb]
    \centering
    \hspace{-5mm}
    \includegraphics[width=0.48\textwidth]{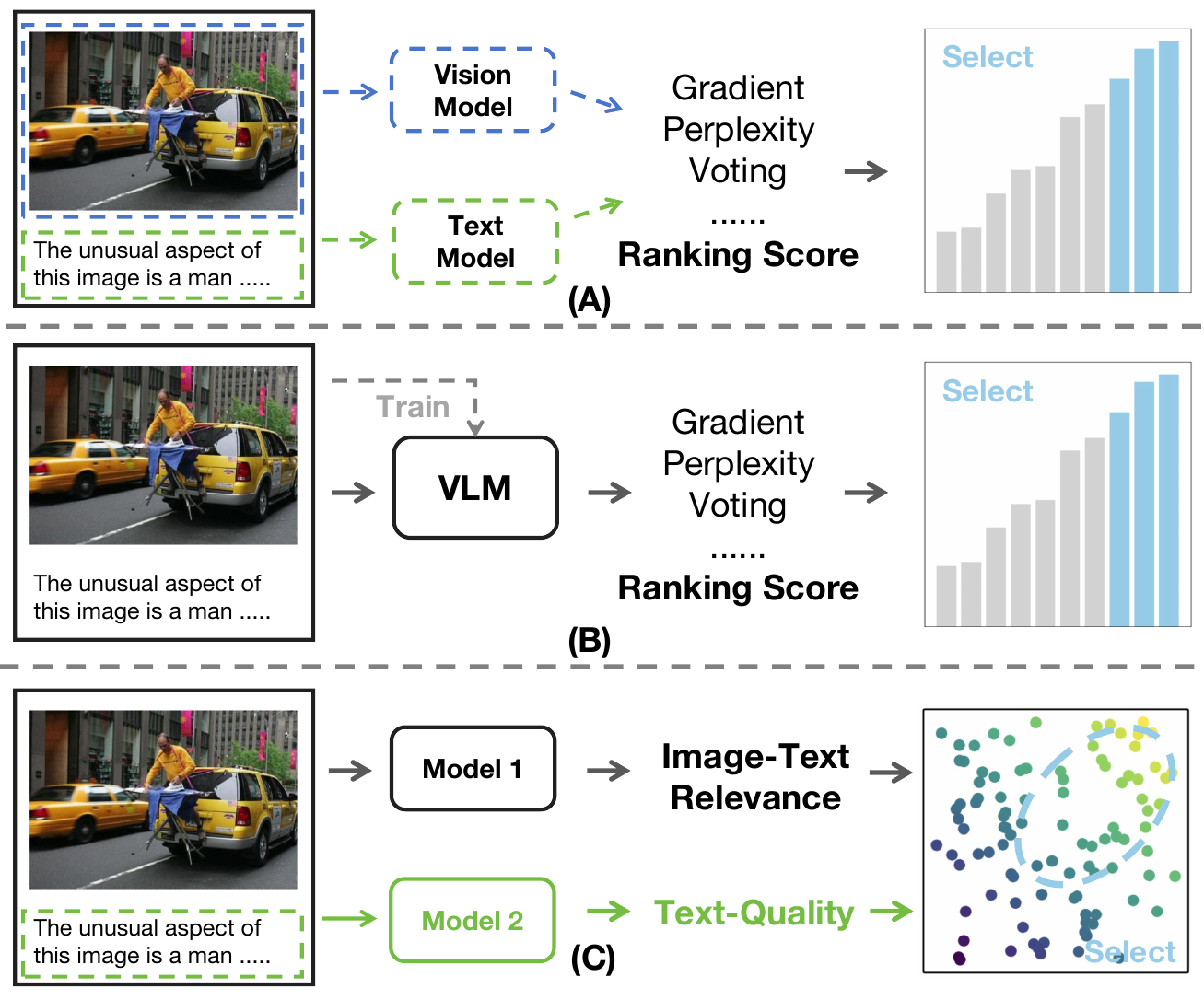} 
    \caption{\textbf{Comparison of data selection methods.}
    \textbf{(A)} Methods based on single-score proxies, often derived from either language or vision signals.
    \textbf{(B)} Methods that use VLMs as quality evaluators, which may suffer from data contamination or prior exposure when the evaluator has been trained on overlapping corpora.
    \textbf{(C)} Our method combines off-the-shelf pretrained models without requiring task-specific training on the target dataset.}
    \label{fig:sample_image001}  
    \vspace{-3mm}
\end{figure}



Large Vision-Language Models (LVLMs) have made significant progress in tasks such as image captioning, visual question answering, and instruction following~\cite{zhu2023minigpt4, dai2023instructblip,bai2023qwen,openai2023gpt4,mishra2019ocr,gemini,xai2024grok,wu2026vision,wang2025infinity}. These models are typically trained in two stages: the first stage performs large-scale image-text pretraining to establish basic vision-language alignment, and the second stage—Visual Instruction Tuning (VIT)—fine-tunes the model on diverse instruction datasets to improve its ability to follow human instructions~\cite{liu2023llava}.


To enhance task generalization, recent work has expanded the scope of VIT by incorporating a broader range of vision-language tasks. However, training on such large-scale instruction data is computationally expensive and often unaffordable for smaller research labs. Moreover, not all data contribute equally to downstream performance. Prior studies have shown that carefully selected subsets can match or even exceed full-dataset training while significantly reducing cost~\cite{zhou2023lima,coincide,wu2024icons}. This motivates a central question: \textit{how can we identify the most valuable data for visual instruction tuning}?

As shown in Figure~\ref{fig:sample_image001} (a), existing data selection methods often rely on proxy signals such as loss, perplexity, confidence scores, gradient norms, or similarity-based heuristics~\cite{el2n,self-filter,perplexity}. While these methods are efficient, they often capture only local training dynamics and fail to reflect a sample’s semantic richness or generalization utility. They are also tightly coupled with model training, requiring early-stage loss traces or backpropagation, which increases computational overhead and reduces portability, as illustrated in Figure~\ref{fig:sample_image001} (b)~\cite{cao2023less,wu2024icons,coincide,hessel2021clipscore,self-filter}. 

In this work, we propose to formulate data quality estimation as a language reasoning task. The core idea is to leverage the zero-shot semantic and logical capabilities of pretrained large language models (LLMs), and use carefully designed prompts to guide the model to assess the quality of each instruction sample based on fluency, informativeness, and instruction alignment, purely via forward inference without relying on any training signals or backpropagation~\cite{sachdeva2024train, openai2023gpt4}. Compared to traditional proxy-based approaches that depend on loss, confidence, or similar heuristics, LLMs provide a more global, semantically grounded perspective, offering advantages in terms of cost-efficiency, generalizability, and robustness. While VLMs also possess multimodal capabilities, their training data often overlaps significantly with the target dataset, and their ability to assess linguistic quality is limited. Therefore, we adopt LLMs as a more neutral and reliable evaluator.

However, high-quality scoring alone does not guarantee effective data selection~\cite{wu2024icons,selcon_coreset,gao2023self,xia2024less}. An equally important component is how samples are chosen based on these scores. To this end, we introduce a lightweight weighted sampling strategy that avoids rigid top-$k$ truncation. Instead of selecting only the highest-ranked examples, our method assigns non-zero sampling probabilities across the score spectrum—preserving rare but informative samples from low-density regions, as illustrated in Figure \ref{fig:sample_image001} (c).  This helps maintain data diversity, mitigates selection bias, and improves model robustness during training. This LLM-as-evaluator paradigm, combined with a soft and diversity-aware sampling scheme, enables the construction of a compact yet informative data subset for vision-language instruction tuning. By identifying and retaining the most useful examples without overfitting to dominant patterns, our method significantly improves training efficiency while maintaining or even improving final performance.

We conducted extensive evaluations on general VQA benchmarks and specialized math tasks using LLaVA-1.5-7B and LLaVA-1.5-13B~\cite{llava1.5} as baselines. Remarkably, DOSE retains 96\% of full-data performance on general VQA using only 20\% of the data and even surpasses full-data results on math tasks with the same 20\% subset. DOSE outperforms methods requiring prior exposure to filtered data, demonstrating superior balance across performance, computational cost, cross-domain generalization, and sample diversity.

Our main contributions are:
\vspace{-3mm}
\begin{itemize}[leftmargin=*]
    \item We propose an efficient data selection method that leverages pre-trained, off-the-shelf models to rapidly assess text quality and image–text relevance, significantly reducing data filtering costs.
    \item Extensive experiments demonstrate that our approach achieves an optimal trade-off between selection efficiency and training performance.
    \item Experiments on multimodal math benchmarks validate that our approach generalizes well to specialized domains, where a small fraction of training data achieves performance comparable to the full training set.
\end{itemize}

\begin{table*}[t!]
\centering 
\scriptsize
\resizebox{\textwidth}{!}{ 
\begin{tabular}{p{2cm} p{12cm}}
\toprule
\textbf{Tasks} & \textbf{Examples of Task Templates} \\ 
\midrule   
Original Template & 
\begin{tabular}[l]{@{}p{12cm}@{}}
    \textcolor{blue}{\textbf{Question}}:  " $\langle image \rangle$   What are the colors of the bus in the image? " \\ 
    \textcolor{teal}{\textbf{Answer}}: " The bus in the image is white and red. "
\end{tabular} \\
\midrule  
Scoring Template &  
\begin{tabular}[l]{@{}p{12cm}@{}}
    \textcolor{blue}{\textbf{Question}}: 
    " \#\#\# What are the colors of the bus in the image? The bus in the image is white and red. \#\#\# Does the previous paragraph demarcated within \#\#\#  contain informative signal for visual instruction tuning a vision-language model? An informative data point should be well-formatted, contain usable knowledge of the world, and strictly NOT have any harmful, racist, sexist, etc. content. OPTIONS: -yes -no " \\ 
    \textcolor{teal}{\textbf{Answer}}: " Response: yes"
\end{tabular} \\
\bottomrule
\end{tabular} 
}
\vspace{-2mm}
\caption{Task template examples.  "Original Template" represents the original format of the data, while "Scoring Template" represents the format used to assist in evaluating the quality of the text within the data.  $\langle image \rangle$ indicates that the original data contains corresponding image information; in the scoring template, we only assess the quality of the textual information, so this token is omitted.}
\label{tab:task_template_examples} 
\end{table*}

\section{Related Work}

\subsection{Data Quality Scoring }
 
Quality-score was originally developed for importance sampling but is now widely used in training LLMs. The scoring algorithm evaluates sample importance using various methods, including measuring disagreement rates between models, assessing whether a sample is likely to be "forgotten", "memorized", or "unlearnable", and applying perplexity filtering to prioritize low-perplexity samples while discarding high-perplexity ones ~\cite{forgetting_events,chitta2021training,feldman2020neural,mindermann2022prioritized,wenzek2019ccnet,marion2023less,muennighoff2023scaling}. Recent advancements have enabled perplexity estimation through efficient model-based simulators, eliminating the need for full LLM inference ~\cite{simfluence}. Additionally, some approaches select training data by minimizing the distance between the selected data distribution and high-quality sources such as Wikipedia or books. This is often achieved through contrastive classifiers or feature-space matching ~\cite{gpt2, palm2, javaheripi2023phi}. To more effectively assess the comprehensive quality of multimodal image-text data, we introduce the CLIP-Score~\cite{hessel2021clipscore} for evaluating image-text relevance. For textual data, we leverage the reasoning capabilities of instruction-tuned LLMs to directly evaluate sample quality. Specifically, we use the acceptance probability assigned by the LLM to measure the likelihood that a given text is valid and meaningful.


\subsection{Data Selection on Distribution}

Data selection is crucial for improving model training quality and can be divided into two categories: distribution-agnostic filtering and distribution-aware selection. Distribution-agnostic methods focus on the quality of individual samples, typically using thresholds to identify subsets~\cite{fang-etal-2017-learning,chen2025murating}. For example, these methods may detect mismatched text-image pairs or misleading elements in images. Specifically, recent works employ BLIP to identify mismatches between captions and images, and leverage OCR models to filter out images where text is the only feature correlated with the caption~\cite{nguyen2023improving, mahmoud2023sieve, maini2023t}. In contrast, distribution-aware methods select subsets by explicitly modeling the overall data distribution. Classical techniques, such as submodular optimization methods, aim to maximize subset performance under a fixed budget~\cite{wei2015submodularity, raskutti2016statistical}. More recently, a codebook-based approach has been proposed, which replaces traditional models, clusters samples, and selects representative samples from each cluster~\cite{wang2023too}. Our method builds upon these ideas by constructing a joint distribution of image-text relevance and text quality. We carefully analyze the impact of different regions and diversity within this joint distribution on data quality, ultimately selecting the most representative samples for training.

\begin{figure*}[t!]
    \centering
    \hspace{-3mm}
    \includegraphics[width=0.98\textwidth]{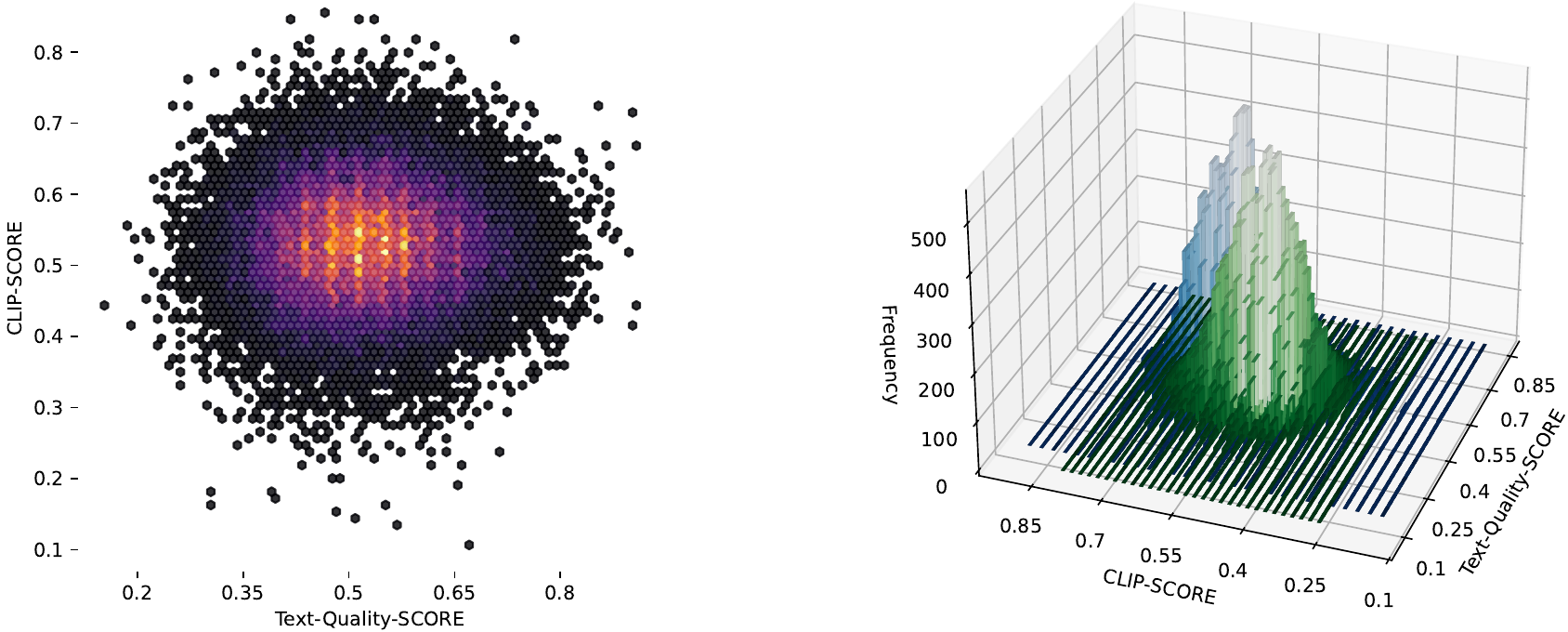}
    \vspace{-2mm}
    \caption{ \textbf{Left: Combined distribution of Text-Quality Score and CLIP Score, with Text-Quality Score on the x-axis and CLIP Score on the y-axis.} Color intensity indicates data density, where brighter colors correspond to higher densities. \textbf{Right: Comparison between the original data distribution and the distribution after applying WRS on 665K samples from LLaVA Stage 2.} The same axes as the left figure are used, with an additional z-axis representing data density. The height along the z-axis corresponds to the density in each region. The blue distribution corresponds to the target distribution $q(x)$, while the green distribution corresponds to the original distribution $p(x)$.
    } 
    \label{fig:sample_imag_com3ssss}  
\end{figure*}


\section{Methodology}

Multimodal data selection mainly focuses on assessment data quality, with existing methods typically assessing text quality and the overall quality of image-text pairs. To obtain a more holistic assessment of multimodal data quality, we combine text-quality estimation with image-text relevance scoring. Existing text quality evaluation methods either introduce bias toward noisy samples with information or face the issue where the evaluation model has already seen the data during training. To address this, we introduce the Text-Quality Score, which leverages the reasoning capabilities of a pre-trained LLM to assess text quality. Additionally, we use the widely adopted CLIP-Score to evaluate the quality of image-text pairs. Meanwhile, selecting data using a static threshold may lead to a loss of diversity and the discarding of valuable edge cases, potentially limiting performance. To address this, we introduce a weighted sampling strategy that balances score-based preference with broader coverage of the empirical score distribution. This approach enables us to select a high-quality subset while maintaining stability and representativeness, improving sample quality while avoiding overly aggressive truncation of long-tail samples.


\subsection{Off-the-Shelf Quality Assessment}

We leverage the reasoning capabilities of pre-trained LLMs and  multimodal language models to evaluate data quality. Inspired by Ask-LLM~\cite{sachdeva2024train}, we prompt the LLM to predict whether an input sample is suitable for fine-tuning a multimodal language model. As illustrated in Table~\ref{tab:task_template_examples}, the prompt asks the LLM to judge whether a sample is suitable for multimodal instruction tuning based on informativeness, coherence, and task relevance. The softmax probability assigned to the “yes” token serves as the \emph{Text-Quality Score} for the sample. In addition, following prior work~\cite{nguyen2023improving,mahmoud2023sieve,maini2023t,fang2023data}, we use CLIP-ViT-B/32~\cite{radford2021learning} to compute the CLIP-Score~\cite{hessel2021clipscore} for assessing the alignment between images and their captions. The CLIP model projects both images and text into a shared embedding space, and the cosine similarity between these embeddings quantitatively measures the image-text relevance.


\subsection{Weighted Random Sampling}
To effectively select high-quality and diverse samples, we define a target distribution $q(x)$ and a Gaussian reference distribution $p(x)$. The latter serves as a smooth approximation to the empirical score distribution.  As shown in Figure~\ref{fig:sample_imag_com3ssss}, the new distribution $q(x)$ shifts density toward high-quality regions while avoiding the over-dominance of high-density middle-score regions and retaining broader coverage over the score spectrum, thereby mitigating the over-representation of moderate-quality, high-density regions. We then perform \textit{Weighted Random Sampling (WRS)} based on $q(x)$, assigning higher sampling probabilities to desirable samples. This strategy biases selection toward higher-quality samples while retaining stochasticity and broader score-space coverage.


\paragraph{Sampling Procedure.}
We begin by computing the statistical properties of the score distribution, including the mean $ \mu_{\text{data}} $ and standard deviation $ \sigma_{\text{data}} $. To obtain a smooth estimate of the data distribution, we apply Kernel Density Estimation (KDE):

\vspace{-7mm}
\begin{equation}
KDE(x) = \frac{1}{N h} \sum_{i=1}^N K\!\left(\frac{x - x_i}{h}\right),
\end{equation}

\noindent
where \(K(\cdot)\) is the Gaussian kernel, \(N\) is the number of samples, and \(h\) is the bandwidth. We then apply KDE to the filtered data and identify the principal mode of the distribution:

\vspace{-7mm}
\begin{equation}
\mu_{\text{peak\_kde}} = \arg\max_{x \in [x_{\min},\,x_{\max}]} KDE(x).
\end{equation}

Next, we get the maximum score as:
\vspace{-3mm}
\begin{equation}
x_{max} = \max  x_i,
\end{equation}



To determine a robust target center, we combine two complementary indicators: the KDE mode $\mu_{\text{peak\_kde}}$, which captures the most representative high-density region of the empirical score distribution, and $x_{max}$, which corresponds to the highest score and serves as a proxy for the highest achievable sample quality. Averaging these two values balances the typical quality level with the upper bound of acceptable sample quality, while avoiding bias introduced by means or medians in imbalanced data. The final target center is defined as:

\vspace{-3mm}
\begin{equation}
\mu_{\text{peak\_wrs}} = \frac{\mu_{\text{peak\_kde}} + x_{max}}{2}.
\end{equation}

Based on $\mu_{\mathrm{peak\_wrs}}$, $q(x)$ and $p(x)$ can be expressed as Gaussian distributions centered at $\mu_{\mathrm{peak\_wrs}}$ and $\mu_{\mathrm{peak\_kde}}$, respectively. Their probability density functions are given as follows:

\vspace{-6mm}
\begin{equation}
\begin{split}
q(x) &= \mathcal{N}\bigl(x;\,\mu_{peak\_wrs},\,\sigma_{\text{data}}\bigr), \\
p(x) &= \mathcal{N}\bigl(x;\,\mu_{peak\_kde},\,\sigma_{\text{data}}\bigr).
\end{split}
\end{equation}

\noindent
To perform WRS, we calculate the weight for each data point $ x_i $ as the ratio of the probability density under the target distribution to that under the original distribution:

\vspace{-5mm}
\begin{equation}
w_i = \frac{q(x_i)}{p(x_i) + \epsilon},
\end{equation}
where $\epsilon = 10^{-10}$ is a small constant added to avoid division by zero. Subsequently, we normalize the weights:

\vspace{-7mm}
\begin{equation}
w_i' = \frac{w_i}{\sum_{j=1}^N w_j}.
\end{equation}

Finally, based on the normalized weights $ w_i' $, we perform weighted random sampling to select $ M $ samples (without replacement) from the original data:

\vspace{-7mm}
\begin{equation}
S_x = \{ x_{i_1}, x_{i_2}, \dots, x_{i_M} \},
\end{equation}

where $ i_k $ are indices randomly drawn according to the weights $ w_i' $. Through these steps, we generate a new sample set $ S_x $ that better aligns with the characteristics of the target distribution $ q(x) $. Also, based on the Image-Text Relevance Scores ($y_i$), we can apply the same sampling strategy to obtain the corresponding subset $ S_y $ :

\vspace{-3mm} 
\begin{equation}
S_y = \{ y_{i_1}, y_{i_2}, \dots, y_{i_M} \},
\end{equation}

\begin{table*}[t!]
    \centering
    \small
    \resizebox{\textwidth}{!}{
        \begin{tabular}{l |c c c c c c c c c |c}
             \toprule
             {\textbf{Method}} & {\textbf{VQAv2}} & {\textbf{GQA}} & {\textbf{VizWiz}} & {\textbf{SQA-I}} & {\textbf{TextVQA}} & {\textbf{MME}} & \multicolumn{2}{c}{\textbf{MMBench}} & {\textbf{LLaVA-W}} & {\textbf{Rel. (\%)}}\\
             & & & & & & & {\textbf{en}} & {\textbf{cn}} & {\textbf{Bench}} & \\
             \midrule
             Full &
             79.1 & 63.0 & 47.8 & 68.4 & 58.2 & 1476.9 & 66.1 & 58.9 & 67.9 & 100\\

             \midrule
            \multicolumn{10}{l}{\textit{Full Data Used before Selection}} \\
            COINCIDE &
            76.5 & 59.8 & 46.8 & 69.2 & 55.6 & 1495.6 & 63.1 & 54.5 & 67.3 & 97.4 \\
            ICONS  &
            76.3 & 60.7 & 50.1 & 70.8 & 55.6 & 1485.7 & 63.1 & 55.8 & 66.1 & 98.6 \\
             \cmidrule{0-10}
             \multicolumn{10}{l}{\textit{Partial Data Used before Selection}} \\
             
              Random &
             75.7 & 57.6 & 44.7 & 66.5 & 54.2 & 1389.0 & 62.2 & 54.8 & 65.0 & 94.5 \\
            CLIP-Score &
             73.4 & 51.4 & 43.0 & 65.0 & \textbf{54.7} & 1331.6 & 55.2 & 52.0 & 66.2 & 91.2\\
              EL2N &
             76.2 & 58.7 & 43.7 & 65.5 & 53.0 & 1439.5 & 53.2 & 47.4 & 64.9 & 92.0\\
              Perplexity &
             75.8 & 57.0 & 47.8 & 65.1 & 52.8 & 1341.4 & 52.0 & 45.8 & 68.3 & 91.6\\
                SemDeDup  &
             74.2 & 54.5 & 46.9 & 65.8 & 55.5 & 1376.9 & 52.2 & 48.5 & \textbf{70.0} & 92.6\\
                D2-Pruning &
             73.0 & 58.4 & 41.9 & \textbf{69.3} & 51.8 & 1391.2 & \textbf{65.7} & \textbf{57.6} & 63.9 & 94.8\\
                Self-Sup  &
             74.9 & 59.5 & 46.0 & 67.8 & 49.3 & 1335.9 & 61.4 & 53.8 & 63.3 & 93.4\\
                Self-Filter  &
             73.7 & 58.3 & \textbf{53.2} & 61.4 & 52.9 & 1306.2 & 48.8 & 45.3 & 64.9 & 90.9\\

            \midrule
            Ours  & \textbf{77.3} & \textbf{58.7} & 46.5 & 67.2 & 54.4 & \textbf{1462.2} & 62.5 & 54.8  & 65.8 & \textbf{96.0}  \\ 
            \bottomrule
        \end{tabular}
    }
\vspace{-2mm}
\caption{Comparisons with baseline methods, with all models trained on 20\% of the full training data and subsets selected by different methods. The best results among methods that do not access the full training data before selection are shown in \textbf{bold}.}
\label{tab:main}
\end{table*}

\paragraph{Combined Sampling.}


Once the positions of all data points are determined in a two-dimensional coordinate space, where each point is defined by $x_i$ representing text quality and $y_i$ representing image-text relevance, we perform score-guided sampling along each dimension and retain samples that are favored by both criteria. This distribution reveals patterns in the data, enabling us to analyze and compare the data distribution before and after sampling. Based on this distribution, we design a sampling strategy that prioritizes regions with both high densities and favorable characteristics in terms of $x_i$ and $y_i$. Specifically, we define two index sets, $I_x$ and $I_y$, which contain the indices of samples selected along the $x_i$ and $y_i$ dimensions, respectively. The final sampled set is obtained by taking the intersection of these two index sets.

\vspace{-7mm}
\begin{equation}
    \textit{DOSE} = \{(x_i, y_i) \mid (x_i, y_i) \in S_x \cap S_y \}.
\end{equation}

This approach ensures that the sampled points not only reflect the underlying data distribution but also align with preferred ranges for text quality and image-text relevance.

\section{Experiments}

We evaluate the effectiveness of our proposed data selection method, DOSE, on a range of visual instruction tuning (VIT) tasks. Our goal is to assess whether DOSE can identify high-value training subsets that achieve competitive or superior performance to baseline methods under a fixed data budget. This section describes our implementation, experimental setup, comparisons against baselines, and efficiency analysis.

\paragraph{Implementation Details.}

All experiments are conducted using the LLaVA-1.5 architecture, focusing on the second-stage VIT process where multimodal instruction-following models are trained on image–text–instruction triples. For the main experiments, we adopt the 7B version of LLaVA-1.5 and apply DOSE to select a 20\% subset (665K samples) from the official VIT dataset, while keeping the pretrained vision and language encoders frozen and retraining only the instruction-tuning stage. Each training sample is assigned two quality scores: a text quality score predicted by Vicuna-7B~\cite{vicuna} and an image–text alignment score computed with CLIP-Score~\cite{hessel2021clipscore}. These scores define a joint 2D distribution in which each point represents a sample’s textual and visual informativeness (as illustrated in Figure~\ref{fig:sample_imag_com3ssss} in the Appendix). We then perform Weighted Reservoir Sampling (WRS) over this 2D space to construct the final training subset, prioritizing samples that are strong in both modalities. To further assess downstream generalization, we also apply DOSE to the MathV360k dataset~\cite{shi-etal-2024-math} and fine-tune LLaVA-1.5-13B; this experiment serves as a case study and is not included in leaderboard comparisons.

\paragraph{Experimental Setup.}

We evaluate DOSE on nine diverse VIT benchmarks: VQAv2, GQA, VizWiz, SQA-I, TextVQA, POPE, MME, MMBench (English), and LLaVA-W. Complete dataset details are provided in the Appendix. Each model is trained using only 20\% of the available VIT data. We report absolute scores using each benchmark’s official evaluation metric, and compute Relative Performance (Rel.\%) by normalizing against the score achieved using the full 100\% VIT training set. To contextualize these results, we also compare DOSE against two categories of data selection methods:

\noindent
\textit{Seen-data selectors}, such as ICONS~\cite{wu2024icons} and COINCIDE~\cite{coincide}, which rely on full-data finetuning to score or cluster training samples.

\noindent
\textit{Unseen-data selectors}, which operate without accessing the full dataset and include Random sampling, CLIP-Score, EL2N~\cite{el2n}, Perplexity~\cite{perplexity}, SemDeDup~\cite{abbas2023semdedup}, D2-Pruning~\cite{d2}, Self-Sup~\cite{selfsup}, and Self-Filter~\cite{self-filter}.

\paragraph{Benchmarks.}

GQA~\cite{GQA2019}, which focuses on reasoning about visual attributes like color and shape, and VQA-v2~\cite{goyal2017making}, which assesses broader visual reasoning. MME~\cite{fu2024mmecomprehensiveevaluationbenchmark} evaluates both perceptual abilities and cognitive reasoning, while TextVQA~\cite{singh2019towards} tests OCR-based reasoning. POPE~\cite{li2023evaluatingobjecthallucinationlarge} addresses object hallucination, assessing models’ ability to avoid generating non-existent objects. VizWiz~\cite{vizwiz} focuses on basic visual reasoning for users who are blind, and ScienceQA~\cite{lu2022learn} evaluates knowledge-grounded question answering. Together, these benchmarks provide a comprehensive test of reasoning, perception, and understanding. Meanwhile, for the Special VQA task, we use MathVista~\cite{lu2023mathvista}, a benchmark designed to assess mathematical reasoning in visual contexts. It comprises 6,141 questions from various datasets and covers categories such as FQA, GPS, MWP, TQA, and VQA. With a focus on arithmetic, algebra, and logic, MathVista includes a diverse range of image types, making it an essential platform for evaluating models’ capabilities in mathematical reasoning.

\begin{figure*}[tb]
    \centering
    \includegraphics[width=1\textwidth]{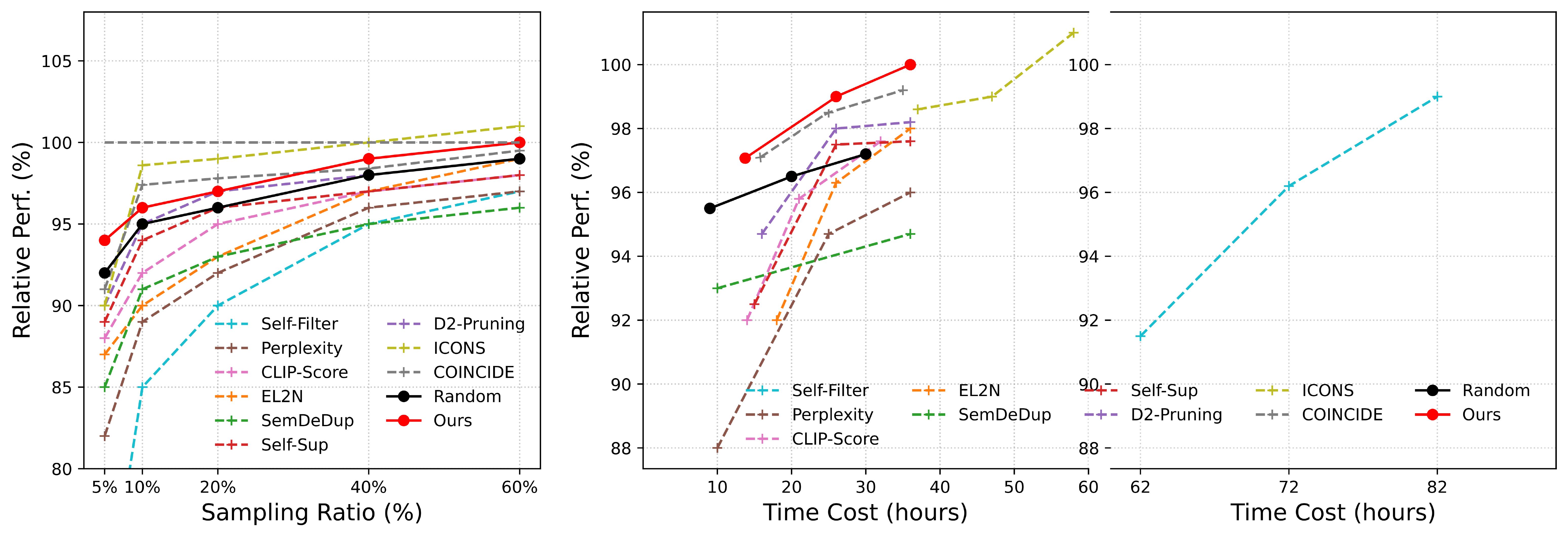}
    \vspace{-7mm}
    \caption{DOSE Data‐Selection Efficiency and Wall‐Clock Time Trade‐Offs. (Left) Average relative performances of all coreset selection techniques at different sampling ratios for the LLaVA-1.5 dataset. (Right) Comparison of coreset selection techniques on average relative performance and wall-clock time cost. The wall-clock time cost includes both the data selection and finetuning of the target VLM. The time cost is measured in hours of running time on a computing node with 4×V100 GPUs. The left panel presents the average relative performance across sampling ratios of 20\%, 40\%, and 60\%. }  
    \label{fig:sample_image3xxx} 
    \vspace{-3mm}
\end{figure*}

\subsection{Main Results and Analysis}

As shown in Table~\ref{tab:baseline3}, DOSE achieves the highest average relative performance (96.0\%) among all unseen-data baselines. It improves upon D2-Pruning (94.8\%) and Self-Filter (93.2\%), and reduces the performance gap to full-data methods such as ICONS (98.6\%) and COINCIDE (97.4\%) to less than 3 percentage points. DOSE consistently outperforms Random across all nine benchmarks (e.g., GQA: 58.6 vs 57.6; TextVQA: 54.4 vs 54.2), and is competitive with or better than more complex selection methods.

While DOSE does not match the top-line scores of full-data selectors, this is expected: ICONS and COINCIDE rely on full-data supervision and exploit task-specific model feedback. In contrast, DOSE requires no additional training, no full-dataset traversal, and operates entirely on pretrained model scores. This leads to significantly lower computational cost and stronger generality, making it particularly suitable for scalable or resource-limited settings.

\paragraph{Different Selection Ratio.}

As shown in Figure \ref{fig:sample_image3xxx} (Left), we compare DOSE (red solid line with circles) against ten baselines—Random (black), Perplexity~\cite{perplexity}, CLIP-Score~\cite{hessel2021clipscore}, EL2N~\cite{el2n}, SemDeDup~\cite{abbas2023semdedup}, Self-Sup~\cite{selfsup}, D2-Pruning~\cite{d2}, COINCIDE~\cite{coincide}, ICONS~\cite{wu2024icons}, and Self-Filter—across sampling ratios from 5 \% to 60 \%. DOSE rapidly climbs to 99 \% Rel. by 40 \% sampling, matching or exceeding all other unseen‐data methods and even approaching the seen‐data ICONS~\cite{wu2024icons} curve at higher ratios.

\paragraph{Efficiency and Performance.}

Among all data selection baselines showen in Figure \ref{fig:sample_image3xxx} (Right), DOSE achieves the largest performance gains among methods that do not rely on prior exposure to the training data, outperforming baselines such as Random, CLIP-Score, EL2N, SemDeDup, Perplexity, Self-Sup, D2-Pruning, and Self-Filter by 1–4 percentage points under identical sampling ratios and time budgets. Even against the two leading seen-data methods, ICONS and COINCIDE, DOSE holds clear advantages. ICONS and COINCIDE both require an expensive full-data fine-tuning pass before sample selection—a cost that would recur for any new dataset yet is omitted from their reported compute comparisons—whereas DOSE skips this phase entirely, relying solely on off-the-shelf pre-trained models for scoring and weighted sampling. As a result, direct comparisons of compute costs are misleading. Moreover, DOSE’s linear-time scoring lets it reach 97.4\% relative performance in ~12 h and 98.5\% in ~22 h, whereas COINCIDE requires ~15 h to reach 97.4\% and ~25 h to reach 98.4\%. ICONS, which lacks a time-optimized pipeline, lags further behind. Finally, DOSE requires no clustering hyperparameters, gradient-influence computations, or extra network training. Its runtime scales linearly with dataset size and is immediately deployable, whereas seen-data methods introduce additional complexity that complicates tuning and extension.

\begin{table*}[ht]
\small 
\centering
\renewcommand{\arraystretch}{1.1}
\setlength{\tabcolsep}{2mm}{
\scalebox{1}{ 
\begin{tabular}{c|cccccc|cccccc|cc} 
\toprule

\multicolumn{1}{c|}{\multirow{2}{*}{\textbf{Size}}} & \multicolumn{13}{c}{\textbf{Math-LLaVA on MathVista}} 

\\ \cmidrule{2-15}
\multicolumn{1}{c|}{} & FQA & GPS & MWP & TQA & VQA & ALG & ARI & GEO & LOG & NUM & SCI & STA & Rel.\% & Aver. \\ 
\midrule
\multicolumn{11}{l}{\textit{Random selection on MathV360K}} \\
\multicolumn{1}{c|}{ 5\%}  & 22.7 & 38.0 & 30.7 & 41.1 & 38.6 & 36.7 & 31.4 & 38.1 & 21.6 & 30.6 & 38.5 & 23.9 & 88.4  & 32.7 \\
\multicolumn{1}{c|}{ 20\%}  & 30.9 & 44.2 & 42.9 & 39.9 & 33.5 & 39.9 & 36.5 & 43.9 & 28.8 & 27.8 & 45.1 & 29.6  & 98.7  & 36.9 \\
\multicolumn{1}{c|}{ 40\%} & 32.3 & 52.4 & 43.0 & 37.3 & 35.2 & 45.6 & 35.7 & 52.3 & 16.2 & 27.8 & 41.9 & 35.9 & 97.6 & 38.0 \\

\midrule
\multicolumn{11}{l}{\textit{DOSE selection on MathV360K}} \\
\multicolumn{1}{c|}{ 5\%}  & 33.4 & 38.9 & 30.1 & 36.1 & 34.1 & 36.3 & 29.5 & 36.8 & 24.3 & 26.4 & 36.1 & 31.9 & 88.4  & 32.8 \\
\multicolumn{1}{c|}{ 10\%}  & 30.5 & 39.9 & 33.9 & 39.9 & 31.8 & 37.4 & 30.0 & 40.2 & 16.2 & 26.7 & 40.2 & 31.9 & 86.8 & 33.2 \\
\multicolumn{1}{c|}{ 20\%} & 33.1 & 45.7 & 45.7 & 42.4 & \textbf{36.9} & 43.1 & 38.5 & 45.2 & \textbf{29.7} & \textbf{31.3} & 41.0 & 35.9 & \textbf{104.8} & 39.1 \\
\multicolumn{1}{c|}{ 40\%} & 32.7 & 49.5 & 47.3 & 43.7 & 34.6 & 47.0 & 37.1 & 49.4 & 18.9 & 27.8 & 40.2 & 37.5 & 100.4 & 38.8 \\
\multicolumn{1}{c|}{ 65\%} & 30.5 & 49.5 & \textbf{53.8} & 42.4 & 29.1 & 44.8 & 37.4 & 48.5 & 8.1 & 24.3 & 41.9 & 37.5 & 93.1 & 37.3 \\
\multicolumn{1}{c|}{\textbf{ 80\%}} & 32.4 & \textbf{53.4} & 49.5 & \textbf{45.6} & 36.3 & \textbf{48.4} & \textbf{39.4} & \textbf{51.9} & 16.2 & 27.8 & \textbf{46.7} & 38.2 & 103.5 & \textbf{40.5} \\
\midrule
\multicolumn{1}{c|}{100\%$^\dagger$} & \textbf{37.9} & 52.8 & 46.8 & 44.3 & 27.9 & \textbf{48.4} & 33.2 & \textbf{51.9} & 18.9 & 23.6 & 45.1 & \textbf{41.9} & 100 & 39.4\\

\bottomrule
\end{tabular}}
} 
\vspace{-2mm}
\caption{\textbf{Comparison with different data selection scales on domain-specific benchmarks.} 
$^\dagger$ represents our reproduced results of Math-LLaVA-13B.
The best results in all tasks are in bold. MathVista is divided in two ways: task type or mathematical skill, and we report the accuracy under each subset. Rel.\%  keep same setting with general benchmarks, and Aver. means the average score of all tasks.}
\label{tab:baseline3}
\end{table*}

\subsection{Unseen-task Generalization.}

As shown in Table~\ref{tab:baseline3}, we filtered the MathV360K dataset and performed continuous fine-tuning on LLaVA-1.5-13B~\cite{llava1.5} using high-quality subsets of varying proportions. In this process, we strictly adhered to the experimental settings of Math-LLaVA~\cite{shi-etal-2024-math}. Since the evaluation on MathVista requires GPT-3.5~\cite{GPT3} to extract key results, and the performance of different period versions may vary, we reproduced the results of Math-LLaVA as a benchmark for comparison. The experimental results demonstrate that our method achieves performance comparable to Math-LLaVA~\cite{shi-etal-2024-math} when using only 20\% of the high-quality data. Furthermore, when using 80\% of the data, the overall performance of the model improves by 1 percentage point. While DOSE generally performs well, it occasionally underperforms compared to random sampling on tasks like GPS (40\%), TQA (5\%), and VQA (5\%) under small sampling ratios. This is mainly because high-score samples tend to cluster in semantically similar regions, leading to reduced diversity and limited generalization. In contrast, random sampling retains a broader variety of examples, which can be more effective in certain tasks.

\subsection{Ablation Study}

In this section, we conduct ablation experiments by comparing different scoring strategies, score-based sampling strategies, and the fusion of these two strategies. The results are presented in Figure \ref{fig:subfig1}, Figure \ref{fig:subfig3}, and Figure \ref{fig:subfig2} in Appendix.

\begin{figure*}[htb]
    \centering
    \begin{subfigure}[b]{1\textwidth}
        \centering
        \includegraphics[width=1\textwidth]{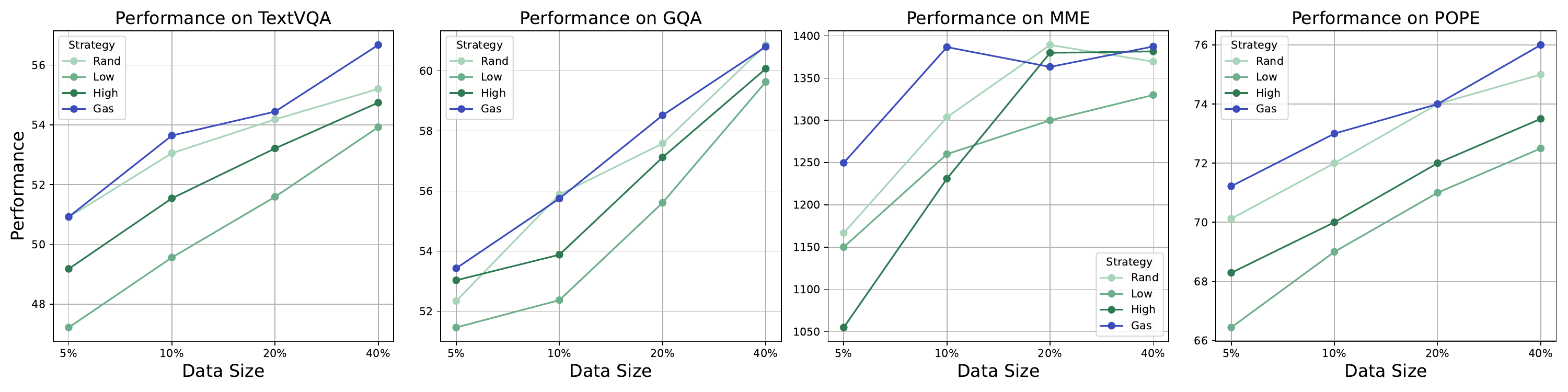}
        \caption{Performance comparison of different strategies based on Text-Quality Score on TextVQA, GQA, MME, and POPE datasets.}
        \label{fig:subfig1}
    \end{subfigure} 
    \begin{subfigure}[b]{1\textwidth}
        \centering
        \includegraphics[width=1\textwidth]{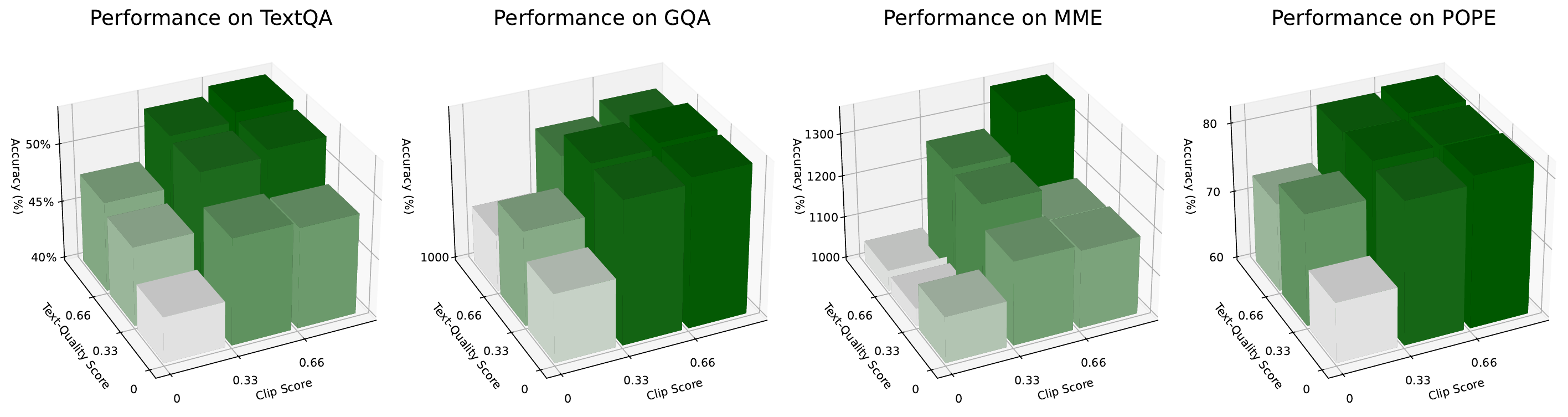}
        \caption{Performance comparison of different regions in the combined distribution, based on CLIP-Score as X-axis and Text-Quality Score as Y-axis.}
        \label{fig:subfig3}
    \end{subfigure}
    \vspace{-7mm} 
    \caption{Overall performance comparisons across different strategies and datasets. (a) and (b) shows the results of the ablation study on the combined distribution, where the height of the columns indicates that taller columns correspond to a darker shade of green. }
    \label{fig:combined_figure} 
    \vspace{-3mm}
\end{figure*}

%

\paragraph{Effectiveness of Filtering Methods}

To evaluate the effectiveness of Text-Quality and CLIP scores independently, we conduct controlled experiments in Stage 2 of the LLaVA training pipeline, as shown in Figure~\ref{fig:subfig1}. We compare four sampling strategies using the Text-Quality Score: \textit{Rand} (random sampling), \textit{High} (top-scoring filtering), \textit{Low} (low-scoring filtering), and \textit{Gas} (Gaussian-based weighted random sampling that balances quality and diversity). Overall, the \textit{High} strategy outperforms \textit{Low}, demonstrating the validity of the Text-Quality score in assessing data quality.  We further observe in Figure~\ref{fig:subfig1} that at a 40\% sampling ratio, \textit{Rand} surpasses \textit{High} on several benchmarks. As discussed in Section~4.3, score-based filtering tends to concentrate on samples with similar language and structure, reducing task diversity and generalization. In contrast, random sampling naturally preserves variation in task types and styles, sometimes yielding better performance. These findings highlight the motivation behind our proposed DOSE framework: combining quality-driven scoring with diversity-aware sampling to achieve a better trade-off. The \textit{Gas} strategy, which embodies this principle, consistently outperforms \textit{Rand}, confirming the effectiveness of our data selection method.

In our evaluation of image-text relevance, shown in Figure \ref{fig:subfig2}, we compared four sampling strategies using the CLIP Score. The results revealed that the “Gas” strategy significantly outperformed the others. This suggests that as the filtering ratio decreases, data quality differences become more noticeable, making it suitable for large datasets with low usage needs. However, as the dataset size grows, the differences in quality between filtered and unfiltered data become smaller. We also found that in the GQA task, the data filtered by CLIP Score did not show significant advantages, likely because the original data already had strong image-text relevance. Such findings underscore the rationale for DOSE, which unifies quality-based and diversity-aware signals to overcome the weaknesses of single-score filtering.

\paragraph{Effectiveness of Combined Sampling}

As shown in Figure \ref{fig:subfig3}, we identified 9 candidate regions based on the original data distribution. These regions represent clusters of data, reflecting the similarities and differences among samples. To create the combined distribution sampling data, we randomly sampled 5\% of the overall data from each candidate region. This method ensures diversity in the samples while effectively capturing the underlying structure of the data. After constructing the combined distribution sampling data, we trained the model using the same settings as the single-method approach and tested it on several datasets, including TextQA~\cite{textvqa}, GQA~\cite{gqa}, POPE~\cite{pope}, and MME~\cite{mme}. And, the performance results are shown in Figure \ref{fig:subfig3}, which indicate that in the upper right area—where both CLIP and Text-Quality Score are high—the model generally performs better. This suggests that in general task, the combination of the two sampling methods can effectively select data that helps improve the model's performance. By using this combined sampling method based on the distribution, we enhance the representativeness and quality of the data, thereby improving the model's training efficiency.


\section{Conclusion}  

In this work, we propose DOSE, an efficient and practical data selection method for multimodal instruction tuning. DOSE leverages off-the-shelf models to evaluate text quality and image-text alignment separately, then combines these scores into a unified quality-alignment distribution for adaptive weighted random sampling. This approach preserves data diversity while identifying the most informative samples. Our experimental evaluation demonstrates DOSE's effectiveness across multiple dimensions. On both general VQA tasks and specialized math benchmarks, DOSE achieves comparable performance to full-dataset training using only 20\% of the data, and surpasses full-dataset results when using 40\% to 80\% subsets. Crucially, DOSE outperforms existing unseen-data selection strategies in both effectiveness and computational efficiency, while operating entirely at inference time without requiring fine-tuning or additional training. These results underscore the critical importance of high-quality data selection in multimodal learning and establish DOSE as a scalable, practical solution for resource-constrained environments.

\section{Limitations}  

While our method demonstrates strong performance and high efficiency, our study is constrained by the experimental cost and a limited exploration budget. We evaluated only an array of sampling ratios and primarily tested our method on LLaVA-1.5 models (7B \& 13B), without assessing more fine-grained sampling ratios or more types of models. As a result, the generality of DOSE across additional sampling ratios and diverse architectures remains to be validated in future work.


\nocite{Ando2005,andrew2007scalable,rasooli-tetrault-2015}

\bibliography{custom}


\appendix

\section{Filtering Details}

We applied the DBSCAN algorithm to filter out anomalous noise from the candidate data.
This subset accounted for approximately 0.01\% of the entire dataset, corresponding to a total of 63 samples.

\begin{table*}[t]
\centering
\renewcommand{\arraystretch}{1.1}
\resizebox{\textwidth}{!}{
\begin{tabular}{llccccccccccccc}
\toprule
Method & Sampling & FQA & GPS & MWP & TQA & VQA & ALG & ARI & GEO & LOG & NUM & SCI & STA & Aver. \\
\midrule
- & Rand 20\%  & 30.86 & 44.23 & 32.26 & 39.87 & 33.52 & 39.86 & 27.76 & 43.93 & 24.32 & 27.78 & 45.08 & 29.57 & 34.92 \\
Clip-Score & TopK 20\% & 24.16 & 32.69 & 26.34 & 42.41 & 37.99 & 32.38 & 29.46 & 32.22 & 13.51 & 30.56 & 40.98 & 28.57 & 30.94 \\
Text-Quality & TopK 20\% & 26.39 & 40.87 & 34.95 & 32.28 & 27.93 & 34.52 & 30.03 & 39.75 & 18.92 & 19.44 & 40.16 & 26.58 & 30.99 \\
Perplexity & TopK 20\% & 33.83 & 30.77 & 31.72 & 39.87 & 31.28 & 29.18 & 28.61 & 33.05 & 10.81 & 24.31 & 45.90 & 35.88 & 31.27 \\
Ours  & TopK 20\%  & 22.68 & 37.98 & 30.65 & 41.14 & 38.55 & 36.65 & 31.44 & 38.08 & 21.62 & 30.56 & 38.52 & 23.92 & 32.65 \\


Ours & WRS 20\% & 22.68 & 43.27 & 36.02 & 39.24 & 34.64 & 38.79 & 32.29 & 42.26 & 20.81 & 30.56 & 45.16 & 35.57 & 35.11 \\

\midrule

- & Rand 40\% & 29.37 & 50.48 & 39.78 & 37.34 & 34.08 & 46.62 & 32.29 & 49.79 & 13.51 & 27.78 & 32.79 & 31.56 & 35.45 \\
Clip-Score & WRS 40\% & 34.20 & 39.90 & 34.95 & 43.04 & 32.96 & 37.37 & 31.16 & 39.33 & 21.62 & 23.61 & 47.54 & 37.21 & 35.24 \\
Perplexity & WRS 40\% & 32.34 & 52.40 & 43.01 & 37.34 & 35.20 & 45.55 & 35.69 & 52.30 & 16.22 & 25.00 & 42.62 & 35.22 & 36.74 \\


Ours & WRS 40\% & 33.5 & 47.2 & 41.4 & 36.7 & 34.6 & 38.4 & 34.3 & 45.6 & 18.9 & 33.3 & 45.9 & 35.2 & 37.08 \\

\bottomrule
\end{tabular}
}
\vspace{-1mm}
\caption{Comparison of different sampling strategies on the Math360k benchmark using LLaVA-1.5-7B.}
\label{tab:math360k123}
\vspace{-3mm}
\end{table*}

\section{Result Analysis}

To understand how our proposed data selection strategy enhances training performance and efficiency, we conducted a visualization and analysis of the data used in LLaVA stage 2, consisting of 665k data points. In the left panel of Figure \ref{fig:sample_imag_com3ssss}, we plotted the CLIP-Score and Text-Quality Score for each data point, revealing a significant concentration of data points in the central area. This suggests that the data likely follows a normal distribution in both scores, indicating regions of higher data quality. These insights led us to examine performance variations across different regions, as discussed in Section 4.3. We found that areas with higher concentrations of data points generally correlated with better performance. This understanding drove us to combine these insights with WRS to create a high-quality data subset selection strategy.

We then visualized the distributions resulting from random sampling (light blue) and WRS sampling (light green) in the right panel of Figure \ref{fig:sample_imag_com3ssss}. The WRS sampling distribution shows a pronounced concentration in regions with higher CLIP and Text-Quality Scores, effectively validating our strategy for assessing data quality and demonstrating the benefits of our sampling approach.

\section{Time Cost Analysis}
 
Figure~\ref{fig:sample_image3xxx} presents a joint analysis of model performance and wall-clock cost across different data selection strategies. The left panel reports the average relative performance of each method under varying sampling ratios (20\%, 40\%, 60\%), while the right panel compares the corresponding total wall-clock time, including both data selection and fine-tuning. Each curve comprises three data points representing these ratios.

Although the x-axes differ (sampling ratio vs. total time), the relationship is direct—higher sampling ratios typically incur greater computational cost. This visualization highlights how different methods navigate the trade-off between efficiency and effectiveness. Among the methods, Perplexity-based filtering exhibits the steepest increase in time cost as the sampling ratio grows. This is due to its inherently sequential and non-parallelizable scoring process, which requires token-level log-likelihood computation for every instruction–response pair. Additionally, Perplexity re-evaluates the selected samples from scratch at each ratio, leading to near-linear or worse scaling behavior in wall-clock time. This limits its scalability to large datasets. Consistent with prior works such as ICONS~\cite{wu2024icons} and COINCIDE~\cite{coincide}, we omit the full-data training cost–performance curve in this figure, as the focus is on fixed-ratio comparisons to highlight efficiency gains.

\begin{table}[ht!]
\centering
\resizebox{\linewidth}{!}{
\begin{tabular}{lccc}
\hline
Method & TextVQA & GQA & MME \\
\hline
Rand 20\% & 54.20 & 57.60 & 1389.00 \\
TopK Clip-Score 20\% & 53.46 & 57.06 & 1404.32 \\
WRS Clip-Score 20\% & 54.59 & 57.72 & 1419.42 \\
TopK Perplexity 20\% & 52.80 & 57.00 & 1341.40 \\
WRS Perplexity 20\% & 53.18 & 57.46 & 1404.37 \\
\hline
\end{tabular}
}
\caption{Results comparison across different methods.}
\label{tab:results12}
\end{table}

\section{Sampling Strategy and Data Quality.}

To better understand the individual and combined effects of data quality scoring and sampling strategy, we conduct two complementary sets of experiments. First, we apply a unified WRS framework to existing scoring methods, including CLIP-Score and Perplexity. As shown in Table~\ref{tab:math360k123}, all methods consistently achieve higher performance under WRS compared to their original Top-K counterparts. This observation indicates that WRS serves as a generally effective sampling mechanism, providing stable performance gains regardless of the specific scoring function. The improvement suggests that introducing stochasticity while preserving score-based preference helps mitigate the limitations of deterministic Top-K selection.

Second, we analyze the impact of different scoring methods under a fixed Top-K setting. As shown in Table~\ref{tab:results12}, our proposed scoring method achieves slightly better performance than existing approaches under the same selection strategy. However, we observe that all Top-K based methods, including ours, struggle to consistently outperform random sampling. This indicates that data quality alone is insufficient to guarantee performance gains when diversity is limited.  When combined with WRS, all scoring methods, including ours, consistently surpass random sampling, highlighting the complementary relationship between data quality estimation and diversity-aware sampling. In particular, WRS enables models to effectively utilize high-quality samples while maintaining sufficient coverage of the data distribution, thereby overcoming the performance ceiling observed in Top-K selection.

Overall, these results demonstrate that (1) WRS is a robust and broadly effective sampling framework, (2) our scoring method provides additional gains under controlled selection settings, and (3) the combination of quality-aware scoring and diversity-preserving sampling is critical for achieving consistent performance improvements.

\begin{figure*}[h]
    \centering
    \includegraphics[width=1\textwidth]{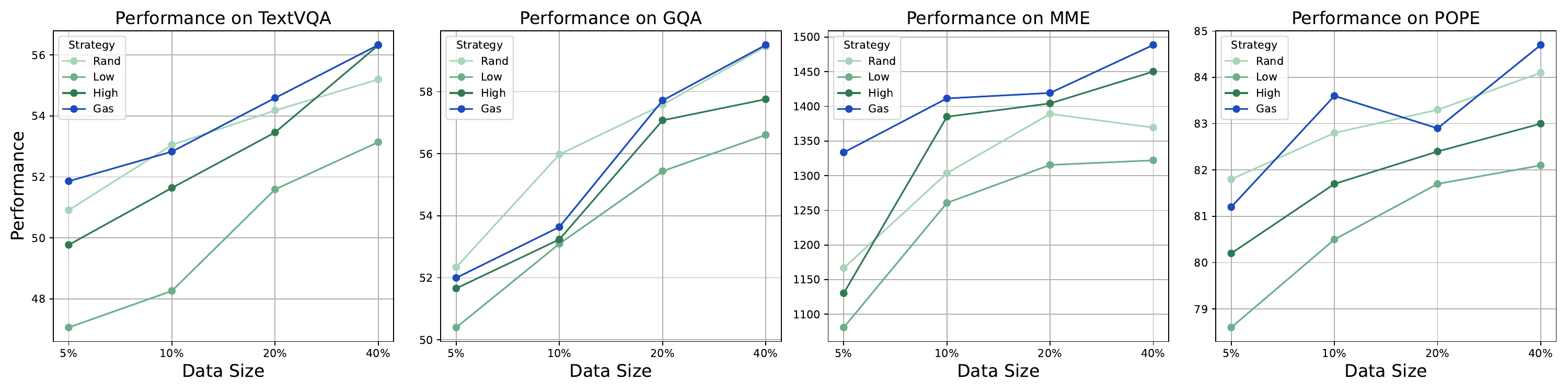}
    \vspace{-3mm}
    \caption{Performance comparison of different strategies based on CLIP-Score on TextVQA, GQA, MME, and POPE datasets.}
    \label{fig:subfig2} 
\end{figure*}

\newpage

\newpage

\newpage

\end{document}